\documentclass[11pt,a4paper]{article}
\usepackage[hyperref]{naaclhlt2018}
\usepackage{times}
\usepackage{latexsym}
\usepackage{graphicx, subfigure}
\usepackage{array}

\usepackage{url}

\aclfinalcopy 

\title{OhioState at SemEval-2018 Task 7:  Exploiting Data Augmentation for Relation Classification in Scientific Papers Using Piecewise Convolutional Neural Networks}

\author{Dushyanta Dhyani \\
  The Ohio State University OH, USA \\
  {\tt dhyani.2@osu.edu} 
  }

\date{}

\begin{document}
\maketitle
\begin{abstract}
  We describe our system for SemEval-2018 Shared Task on Semantic Relation Extraction and Classification in Scientific Papers where we focus on the Classification task. Our simple piecewise convolution neural network (PCNN)  performs decently in an end to end manner. A simple inter-task data augmentation significantly boosts the performance of the model. Our best-performing systems stood 8th out of 20 teams on the classification task on noisy data and 12th out of 28 teams on the classification task on clean data.
\end{abstract}

\section{Introduction}

Relation extraction (RE) and Classification (RC) is an integral component of information extraction systems which aim to extract all the entity pairs and their relation \textit{$\langle e_{1},r,e_{2} \rangle$} from a given text corpora. An alternate formulation of relation extraction task focuses on identifying if a relation exists between a predefined pair of entities, and if yes classify from a given set of class relations. RE finds applications in a variety of domains, ranging from knowledge base construction to semantic parsing and question answering. However, the applicability of existing efforts in relation extraction to scientific text calls for a quantitative and qualitative analysis which is the aim of this shared task.

\section{Related Work}

Existing efforts for RE range from traditional strategies \cite{qian-EtAl:2008:PAPERS,bunescu:nips05,bunescu-mooney:2005:HLTEMNLP,mintz-EtAl:2009:ACLIJCNLP,riedel10modeling} to more recent end to end deep learning based methods \cite{zeng-EtAl:2014:Coling, zeng-EtAl:2015:EMNLP,lin-EtAl:2016:P16-1,wu-bamman-russell:2017:EMNLP2017} that are more suitable in situations where a lot of training data is available. While a majority of efforts in the RE community are specifically focused towards using distantly supervised data and reduce the associated noise, their discussion is not relevant to the current scenario. The most relevant work is that of \cite{zeng-EtAl:2014:Coling} who demonstrated the efficacy of convolution neural networks for relation classification and \cite{zeng-EtAl:2015:EMNLP} who further enhanced the architecture by proposing the piecewise max-pooling strategy. 

\section{Task Description}

\begin{figure*}[t]
    \centering
    \includegraphics[width=0.8\textwidth, height=10cm,keepaspectratio]{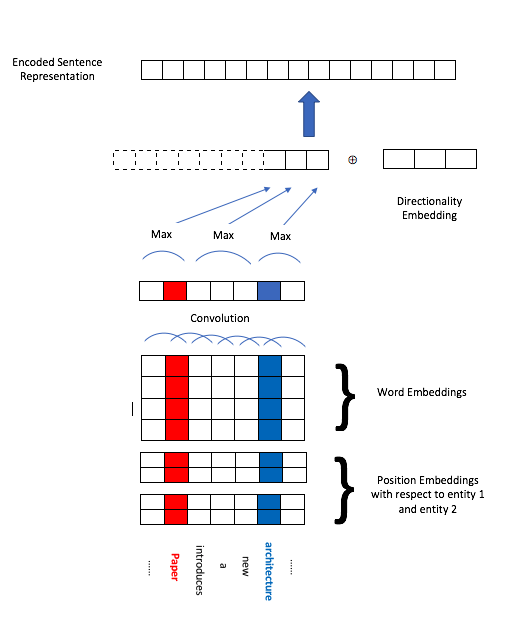}
    \caption{PCNN Encoder with Word, Position and Directionality embeddings.}
    \label{fig:encoder}
\end{figure*}

The semantic relation extraction and classification in scientific papers task \cite{SemEval2018Task7} aims at identifying semantic relations expressed by entity pairs in scientific literature. The contest is further divided into three subtasks, where the first two focus on classification of varying nature of data and the third focuses on extraction task. Since our submitted systems focused only on the classification task, we would from here on discuss mostly about the classification sub-tasks.   

\subsection{Dataset}

\begin{figure}
\centering     
\subfigure[Task 1]{\label{fig:task1}\includegraphics[width=60mm]{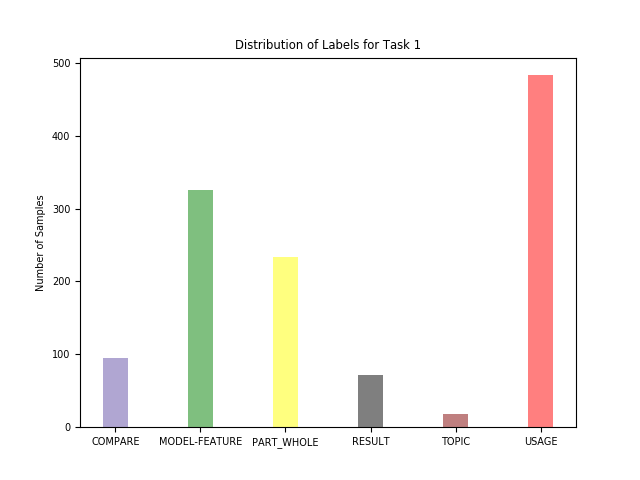}}
\subfigure[Task 2]{\label{fig:task2}\includegraphics[width=60mm]{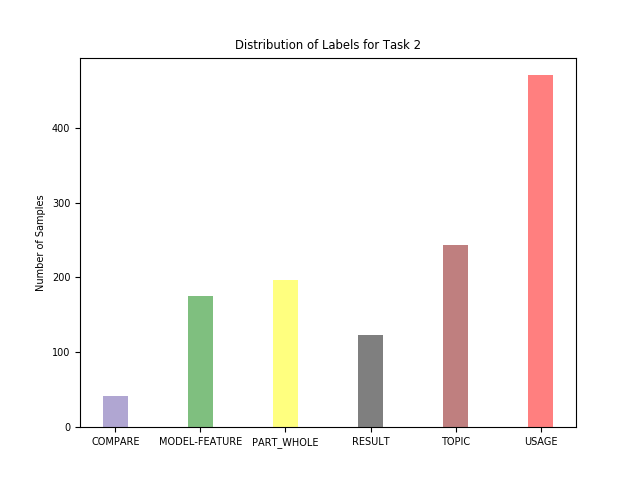}}
\caption{Class Sizes for Task 1 \& Task 2.}
\label{fig:classsize}
\end{figure}

The data contains titles and abstracts of papers from ACL Anthology Corpus where entity mentions are either manually annotated (Subtask 1.1 and Subtask 2) or heuristically (Subtask 1.2) determined. However, the relations are manually annotated across all subtasks. For the classification scenario, we are provided with relevant entities and the directionality of their relation. There are 6 class labels: USAGE, RESULT, MODEL, PART\_WHOLE, TOPIC, COMPARISON. The classes are highly imbalanced in nature as shown in Fig. \ref{fig:classsize}


\subsection{Evaluation}

For both Task 1.1 and 1.2, given that the classes are imbalanced, macro-f1 score is used as the official evaluation metric and thus the metric we use for hyperparameter tuning. For more details, we would refer the reader to the task description paper \cite{SemEval2018Task7}

\section{Methodology}

As shown in Fig. \ref{fig:encoder}, we use the piecewise convolutional encoder proposed by \cite{zeng-EtAl:2015:EMNLP} which encodes the sentence into an embedding space taking into account the context of text around the entities in an end to end manner. The various components of the encoder are described below

\subsection{Preprocessing}

Since the original training dataset provided is annotated using XML tags which can be utilized in a variety of ways, we briefly describe our preprocessing steps. Each text item contains a title of a paper and its abstract. Both the entities for a particular training/testing instance could only either be in the title or in the abstract. While it would be interesting to see the impact of incorporating the effect of paper titles on the entities in abstracts and vice versa, to simplify the architecture, we simply treat titles and abstracts as separate and independent sentences. For the representation of entities, the two most obvious options are to either combine sub-words in an entity using a special character (e.g. \textit{word sense disambiguation} becomes \textit{word\_sense\_disambiguation} ) or to simply use entity head words to represent the starting position of the entity as proposed by \cite{nguyen2015relation}. We chose the latter approach for two reasons: 1) The amount of data is relatively small to learn word embeddings on the data itself 2) The conjoined entity representation as in the former approach would probably not exist in the pre-trained word embeddings and thus would have to be replaced by an unknown token. Finally, we used common text cleaning techniques like removing non-alphanumeric characters, replacing all numbers by a unique token, etc.

\subsection{Word Representation}
Each word in the input is transformed to a static, dense feature representation by looking up a pre-trained word embedding dictionary. We use dependency based word embeddings \cite{levy-goldberg:2014:P14-2} which incorporate long-range dependencies between words and thus generate embeddings that are more functional in nature (than the traditional bag of words based embeddings) which is presumably more suitable to the current task as dependency based features have been shown to be useful for relation extraction \citep{D15-1206,bunescu-mooney:2005:HLTEMNLP}. All words that do not exist in the dictionary are replaced by \textit{UNK} token and initialized randomly.  

\subsection{Position Embedding}

Since convolved representations are position invariant, incorporating positional information using embeddings has been shown to be useful for a variety of task \citep{zeng-EtAl:2014:Coling,P17-1012} when using a convolutional encoder.
We evaluate the distance of each word in the sentence with respect to both \textit{entity 1} and \textit{entity 2} (we limit the values to a maximum distance of \textit{position\_window\_size}). These position values are then projected into a relatively small embedding space using a trainable embedding layer. 

\subsection{Directionality Embedding}

Since the relations are directional in nature, it is important to incorporate the available directionality information in the sentence representation. While this can be implicitly done when using dependency tree base input representation, to incorporate the directionality of the relation exhibited by the two entities ($<e_{1},r,e_{2}>$ or $<e_{2},r,e_{1}>$) we also project the direction information into the embedding space by another embedding layer that is trained along with the entire network.

\subsection{Convolution and Piecewise Max-Pooling}

CNN's have been shown to be good at encoding sentences into vector representations for text classification tasks \cite{kim:2014:EMNLP2014,NIPS2015_5782,NIPS2014_5550,kim2016character} and at the same time also speed up the training and inference time. The word representations and position embeddings are concatenated and fed into a convolution encoder which generates features using varying width of filters. To take into account the context of text around and between the entities in consideration, we then perform a piecewise max-pooling operation as shown in Fig. \ref{fig:encoder}. The input representations (word-embedding $\oplus$ position-embedding) are appropriately padded before the convolution operation to ensure that the convolved features have the same length as the input sentence in order to correctly use entity positions for piecewise max-pooling. These features generated by the PCNN are finally concatenated with the directionality embeddings discussed above to generate the sentence level representation.

\subsection{Regularization, Output and Training}

We use dropout \cite{JMLR:v15:srivastava14a} on the sentence representations with a keep probability of 0.5 as a simple regularization strategy. This is followed by a fully connected layer and a softmax operation for the classification task. We use the standard multi-class cross-entropy loss as our training objective and Adam \cite{kingma2014adam}  for optimization.

\begin{table}[h!]
\centering
\begin{tabular}{|>{\centering\arraybackslash}p{0.6\linewidth}|>{\centering\arraybackslash}p{0.28\linewidth} |} 
 \hline
 \textbf{Parameter} & \textbf{Values}  \\ [0.5ex] 
 \hline
 \textbf{Number of Epochs} & 100,200,400 \\ 
 \textbf{Maximum Sequence Length} & 100,200 \\
 \textbf{Batch Size} & 32,64 \\
 \textbf{Number of Filters} & 32,64,128  \\
 \textbf{Learning Rate} & 0.001, 0.0005 \\  [1ex] 
 \hline
\end{tabular}
\caption{Hyperparameter Values.}
\label{table:paramval}
\end{table}

\section{Experiments}

\begin{table*}[t]
\centering
\begin{tabular}{|c|c|c|c|c|c|} 
 \hline
 \textbf{Task} & \textbf{Data} & \textbf{Epoch} & \textbf{Batch Size} & \textbf{No. of Filters} & \textbf{Macro-F1 Score}  \\ [0.5ex] 
 \hline
1.1 & 1.1 & 200 & 32 & 64 & 35.3 \\  [1ex] 
1.1 & 1.1 + 1.2 & 200 & 64 & 32 & \textbf{48.1}  \\  [1ex] 
1.2 & 1.2 & 200 & 32 & 64 & 64.4 \\  [1ex] 
1.2 & 1.1 + 1.2 & 100 & 64 & 128 & \textbf{74.7}  \\  [1ex] 
 \hline
\end{tabular}
\caption{Results of our best performing systems on the official test set with/without data augmentation.}
\label{table:results}
\end{table*}

\subsection{Data Augmentation}

Deep neural models require significant amount of training data to extract relevant features. While our neural model is relatively shallow, the data size for each of the subtask is also small. As a workaround, we simply mix the data from subtask 1.1 with data from subtask 1.2 which hopefully helps in improving the model's generalizability.

\subsection{Experimental Settings}

While the final training and prediction was performed on the entire training dataset, we use the official validation split provided by contest organizers to perform hyper-parameter tuning. For the data augmentation scenario, however, we also make use of the validation data from the other task. Given that CNN's are fast to train, we easily use grid search to find the optimal combination of a subset of parameters for each task and each data configuration (with or without augmentation) which are listed in Table \ref{table:paramval}. For the remaining parameters, we used standard values as recommended by prior literature as follows: convolution filters of width 3,4 and 5; position and directionality embeddings of size 5; windows size for relative positions from entities was set to 30.

\section{Results}

We report our performance on the classification tasks (Subtask 1.1 and 1.2) according to the official evaluation. While all task settings perform best for a maximum sequence length of 200 and learning rate of 0.001, the rest of the parameters and their corresponding results are listed in Table \ref{table:results}. Even a simple mixing of the two datasets which differ significantly in the nature of tagged entities lead to a significant improvement. Surprisingly though, adding the noisy data to the clean dataset also leads to a 36\% increase in performance. This could be attributed to the fact that while heuristically annotated entities are high-level concepts thus sharing a lot of context with similar concepts, most of the manually annotated entities are full noun phrases, thus adding to the complexity of the task. These results also falsify our initial assumption/expectation of Task 1.1 to be easier.

\section{Conclusion}

We presented a simple end to end model that is fast to train and though does not perform competitively well, makes effective use of additional data for a significant improvement in performance. These results show the effectiveness of mixing/transferring supervision from data coming from a different distribution and thus invites further exploration in semi-supervised/supervised domain adaptation scenarios.    
\bibliography{semeval2018}
\bibliographystyle{acl_natbib}

\end{document}